\documentclass[letterpaper, 10 pt, conference]{ieeeconf}  

\IEEEoverridecommandlockouts                              

\newcommand\egno{\textit{e.g.}}
\newcommand\ieno{\textit{i.e.}}

\usepackage{pifont}
\usepackage{xcolor}

\newcommand{\cmark}{\textcolor{green!60!black}{\checkmark}}
\newcommand{\xmark}{\textcolor{red!70!black}{\ding{55}}}

\usepackage{graphics} 
\usepackage{epsfig} 
\usepackage{mathptmx} 
\usepackage{multirow}
\usepackage{makecell}
\usepackage{booktabs}
\usepackage[table,xcdraw]{xcolor}
\usepackage[normalem]{ulem}
\useunder{\uline}{\ul}{}
\usepackage{hyperref}
\usepackage{color, colortbl}
\usepackage{stfloats}
\usepackage{subfig}
\usepackage{stfloats}
\usepackage[ruled,vlined]{algorithm2e}
\usepackage{float}
\usepackage{amsfonts} 

\let\labelindent\relax
\usepackage{enumitem}

\usepackage{amsmath} 
\title{\LARGE \bf
Token Expand-Merge: Training-Free Token Compression for\\Vision-Language-Action Models
}

\author{Yifan Ye$^{1, *}$,  Jiaqi Ma$^{2, *}$, Jun Cen$^{3}$, Zhihe Lu$^{1,\dagger}$
\thanks{Yifan Ye and Zhihe Lu are with $^{1}$College of Science and Engineering, Hamad Bin Khalifa University, Education City, Doha 24404, Qatar}%
\thanks{Jiaqi Ma is with $^{2}$Mohamed bin Zayed University of Artificial Intelligence, Abu Dhabi 23201, UAE}%
\thanks{Jun Cen is with $^{3}$College of Computer Science and Technology, Zhejiang University, Hangzhou 310058, China}%
\thanks{$^*$ Equal contribution.}
\thanks{$^{\dagger}$ Corresponding Authors, {\tt\small zlu@hbku.edu.qa,}}}

\begin{document}
\maketitle
\begin{abstract}
Vision-Language-Action (VLA) models pretrained on large-scale multimodal datasets have emerged as powerful foundations for robotic perception and control. 
However, their massive scale, often billions of parameters, poses significant challenges for real-time deployment, as inference becomes computationally expensive and latency-sensitive in dynamic environments.
To address this, we propose Token Expand-and-Merge-VLA (TEAM-VLA), a training-free token compression framework that accelerates VLA inference while preserving task performance. 
TEAM-VLA introduces a dynamic token expansion mechanism that identifies and samples additional informative tokens in the spatial vicinity of attention-highlighted regions, enhancing contextual completeness. 
These expanded tokens are then selectively merged in deeper layers under action-aware guidance, effectively reducing redundancy while maintaining semantic coherence. 
By coupling expansion and merging within a single feed-forward pass, TEAM-VLA achieves a balanced trade-off between efficiency and effectiveness, without any retraining or parameter updates.
Extensive experiments on LIBERO benchmark demonstrate that TEAM-VLA consistently improves inference speed while maintaining or even surpassing the task success rate of full VLA models. 
The code is public available on \href{https://github.com/Jasper-aaa/TEAM-VLA}{https://github.com/Jasper-aaa/TEAM-VLA}
\end{abstract}    
\begin{figure}[t]
  \centering
  \resizebox{0.43\textwidth}{0.65\textwidth}{%
    \includegraphics{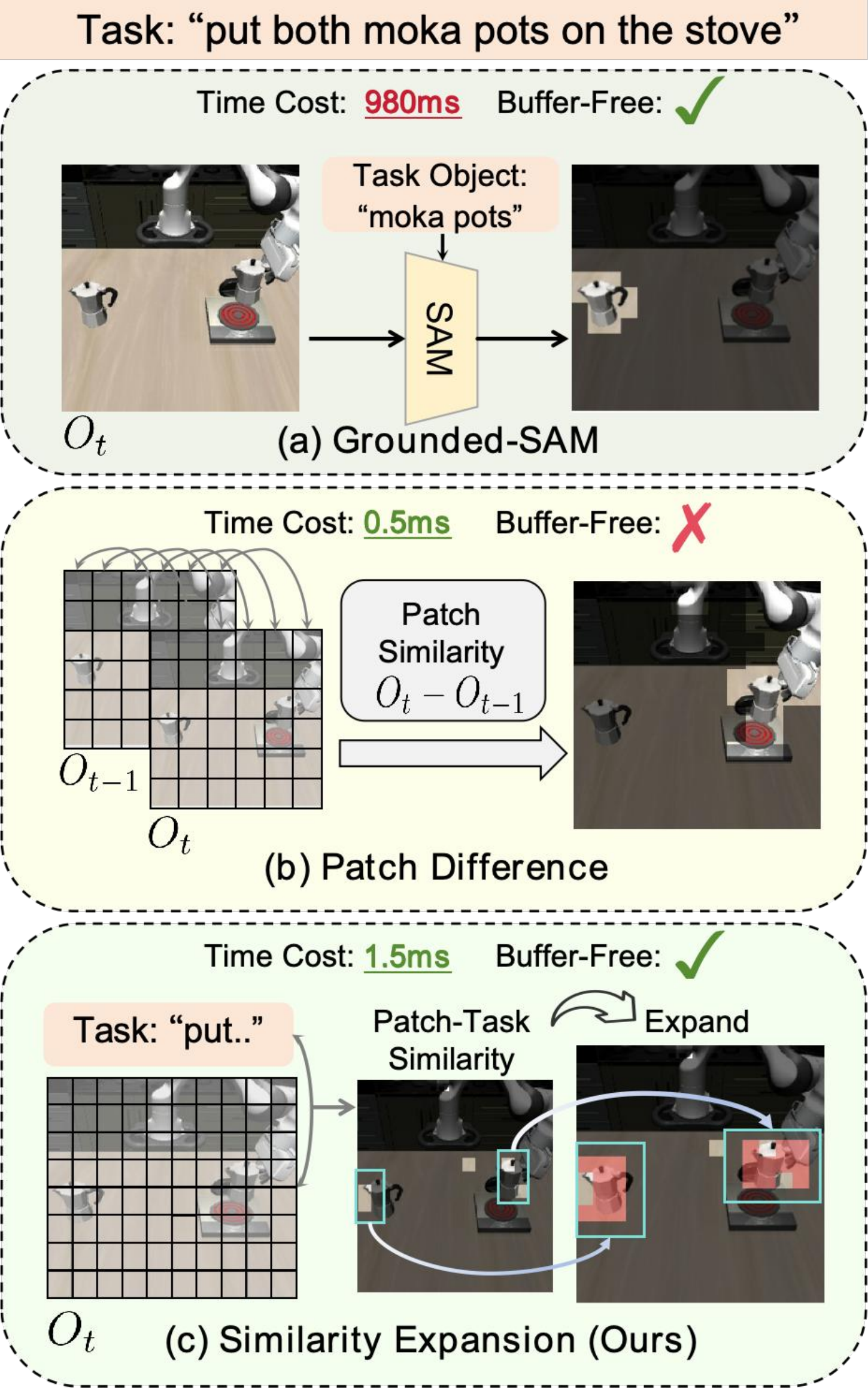}%
  }
  \caption{A visual comparison of three foreground extraction strategies shows that our similarity-driven expansion achieves the most coherent foreground regions while maintaining superior efficiency and zero-buffer overhead.}
  \label{fig:expand}
\end{figure}
\section{Introduction}
Vision–Language–Action (VLA) models have recently demonstrated strong performance across both real-world and simulated robotic control tasks, largely owing to the powerful representation capabilities of Large Vision–Language Models (LVLMs) \cite{openvla,pi0,openvla-oft}. 
While LVLM backbones offer rich multimodal understanding and robust generalization, they also impose substantial computational and memory overhead. 
This significantly limits their practicality in high-frequency, low-latency control scenarios such as real-time manipulation, closed-loop feedback policies, and on-device robotics \cite{vlasurvey,pei2025action,yu2025survey}.
As a result, improving the computational efficiency of VLA models, without sacrificing their action reasoning ability, remains a critical yet underexplored challenge for scalable deployment.


Existing works often mitigate this computational burden through token-level pruning, which accelerates inference by reducing the number of visual tokens processed by the VLA backbone \cite{li2025egoprune,liu2025ttf,jiang2025better,efficientvla,zhang2024sparsevlm}. 
While effective, these approaches typically depend on trainable query mechanisms \cite{jiang2025better} or cross-frame temporal cues \cite{wang2025specprune,xu2025vla,liu2025ttf} to identify and retain salient foreground tokens. 
Such designs often require either additional training or access to previous observations $O_{t-1}$ during inference, which increases system complexity, introduces memory overhead, and may degrade robustness when temporal continuity is unreliable (\egno, abrupt viewpoint changes or partial observability). 
Consequently, a training-free, and temporally independent token compression strategy remains highly desirable for practical VLA deployment.


In this paper, we propose a training-free, observation-only token compression method, named \textbf{T}oken \textbf{E}xpanding \textbf{A}nd \textbf{M}erging for \textbf{V}ision–\textbf{L}anguage–\textbf{A}ction models (TEAM-VLA).
TEAM-VLA identifies foreground-relevant visual tokens purely from the current frame, without relying on trainable queries or temporal buffering. 
Prior works \cite{jiang2025better,wang2025specprune,zhang2024sparsevlm,merging1} suggest that the similarity between projected visual tokens and language embeddings can serve as an indicator of object relevance. 
However, as illustrated in Figure~\ref{fig:expand}, the maximum similarity over all task language tokens yields extremely sparse responses. 
This sparsity arises because task descriptions contain many non-object terms (\egno, ``put'', ``the'', ``on''), each mapping to isolated and semantically irrelevant pixels, while even object-centric phrases (\egno, ``both moka pots'') correspond to only a few token-level anchors. 
This motivates the need to reconstruct the full spatial extent of the underlying objects from these sparse cues. 
To address this, we introduce a Token Expanding mechanism that propagates high-similarity seeds into coherent spatial regions. 
A smoothing convolutional scan is applied to selectively expand linguistically meaningful areas, while noisy or weak-response regions are supplemented through a controlled random expansion to preserve potential foreground candidates.
To maintain overall structural completeness, we further enrich the expanded regions with a small fraction of randomly sampled contextual tokens.

Beyond expansion, we observe that action–text interactions at intermediate layers of the VLA backbone reveal additional visual tokens that encode task-relevant motion cues or spatial structure. Although these tokens may exhibit weaker object relevance, discarding them risks losing important functional information. 
Therefore, TEAM-VLA retains the top-$M$ action–text-responsive tokens and applies a soft bipartite merging mechanism to compress the remaining tokens into semantically aligned groups. 
This ensures that informative yet subtle cues remain preserved in compact form. 
Extensive experiments on the LIBERO benchmark \cite{libero} demonstrate that TEAM-VLA achieves an excellent success–latency trade-off, substantially accelerating inference while maintaining strong execution performance.


\begin{figure*}[t]
  \centering
  \includegraphics[width=\textwidth]{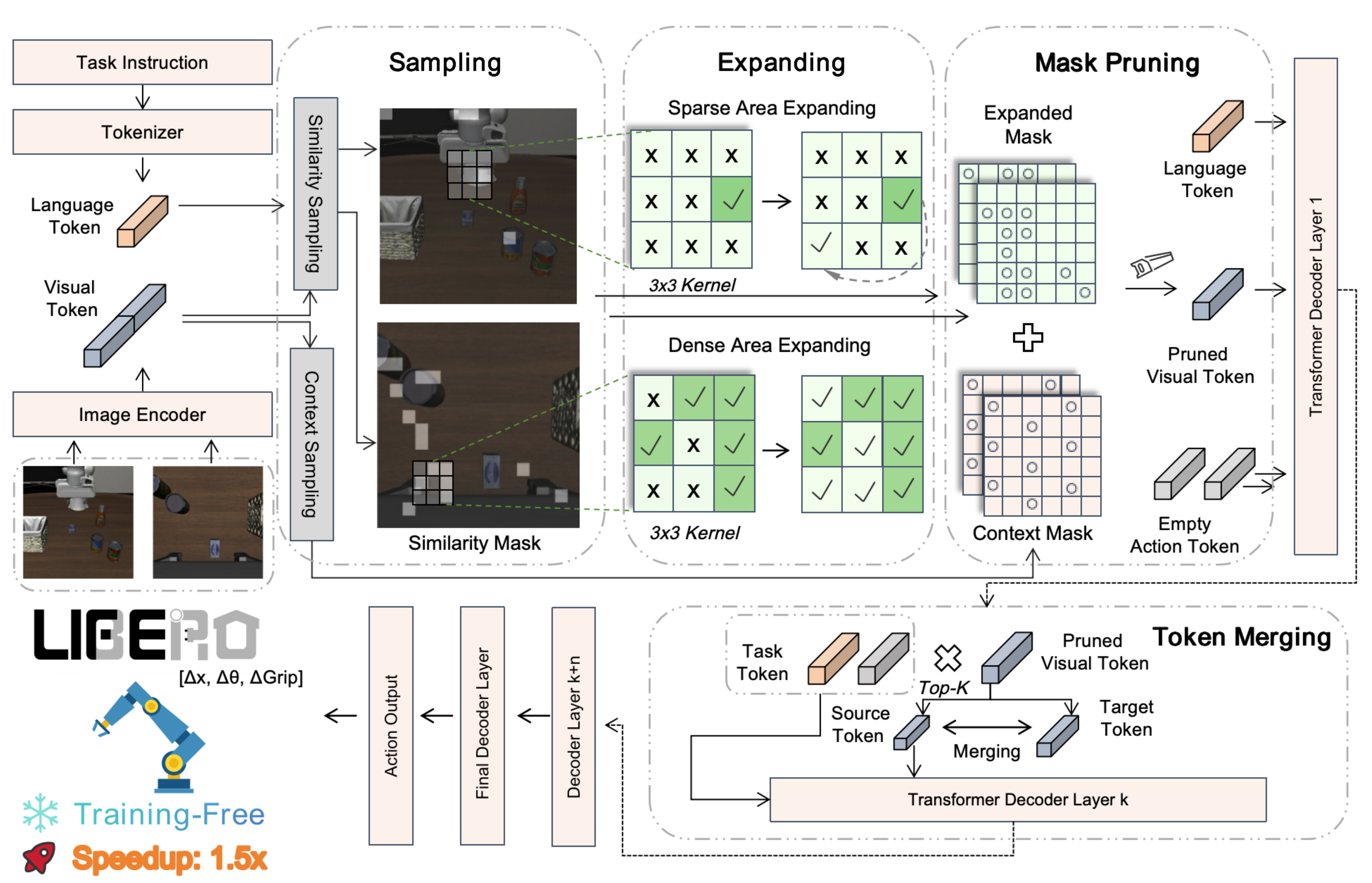} 
  \caption{The overall pipeline of TEAM consists of two stages. First, before tokens enter the language backbone, we perform token pruning. This begins with two complementary sampling steps: (1) similarity-based sampling and (2) context sampling, followed by a spatial expansion module that densifies the sparse similarity map. The expanded mask is then used to remove redundant visual tokens while preserving task-relevant regions. Second, at a middle layer of the backbone, we introduce an action-guided soft bipartite matching module that merges tokens through weighted averaging, effectively compressing deep representations while retaining essential semantic and action-related information.}
  \label{fig:main}
  \vspace{-0.4cm}
\end{figure*}

In summary, our contributions are threefold.
\begin{itemize}
    \item We introduce TEAM-VLA, a fully training-free and observation-only token compression framework for Vision–Language–Action models, requiring no additional supervision, past-buffering, or model retraining.
    \item We develop a foreground-aware token compression pipeline that combines (i) a fast similarity-driven \emph{Token Expanding} module to reconstruct dense foreground regions from sparse vision–language cues, and (ii) an action-guided soft-bipartite \emph{Token Merging} mechanism that compresses deeper-layer tokens while preserving essential semantic structure.
    \item Extensive experiments on the LIBERO benchmark demonstrate that TEAM-VLA achieves an excellent trade-off between token reduction and action success rate, offering a practical acceleration strategy for VLA deployment.
\end{itemize}

\section{Related Work}
\label{sec:related}

\subsection{Vision-Language-Action Models}

Vision–Language–Action (VLA) models extend the capabilities of large vision–language models (VLMs) \cite{vlmsurvey,qwen,minigpt} by integrating a visual–language encoder with an action-generation head \cite{vlasurvey,pi0,li2024cogact,hung2025nora}. 
This unified architecture enables robots to perceive, interpret, and act within diverse environments, yielding strong performance in both simulation and real-world deployments \cite{vlasurvey2,vlasurvey3}. 
Early efforts in this direction include the RT-series models \cite{rt1,rt2}, which combine pretrained VLMs with large-scale robot demonstrations, showcasing the potential of VLA systems to achieve robust semantic understanding and reliable control.
Subsequent works have further advanced this paradigm: $\pi0$ and OpenVLA \cite{openvla} demonstrate remarkable zero-shot generalization by training on extensive real-world datasets, while OpenVLA-OFT \cite{openvla-oft} introduces techniques such as action chunking and parallel decoding to significantly boost execution accuracy—achieving over 95\% on the LIBERO benchmark \cite{libero}. 
These developments highlight the growing maturity of VLA architectures and underline the importance of efficient, scalable methods for real-world deployment.


\subsection{Token Compression for VLA}

Large VLMs (LVLMs) provide powerful semantic grounding for Vision–Language–Action (VLA) systems, yet their heavy Transformer backbones introduce substantial computational overhead when processing long visual token sequences \cite{liang2025large,yu2025survey}. 
As a result, pruning redundant visual tokens has emerged as an effective strategy to improve efficiency in multimodal LVLMs \cite{zhang2024sparsevlm,FastV,vlasurvey3}. 
The core challenge in token pruning is to distinguish informative tokens from unimportant ones \cite{li2025egoprune,kim2022learned,merging1,merging2}. 
Prior work frequently leverages text–image cross-attention maps to determine which patch tokens to retain \cite{zhang2024sparsevlm,kim2022learned,merging}, an approach that works well in general LVLMs because they contain rich and diverse language embeddings. 
However, in robot manipulation, language tokens are typically limited to the task instruction and thus provide sparse guidance. 
This sparsity causes cross-attention–based pruning to degrade, often requiring additional learnable modules or adapters to compensate for the insufficient attention signals \cite{jiang2025better,yu2025survey}.

Beyond static cross-attention pruning, several studies exploit the streaming nature of robotic perception by identifying regions of interest through inter-frame changes \cite{wang2025specprune,liu2025ttf,xu2025vla,li2025egoprune}.
Methods such as VLA-Cache \cite{xu2025vla} and SpecPrune-VLA \cite{wang2025specprune} adopt a two-frame comparison strategy to select dynamic or foreground-relevant tokens, effectively preserving motion cues. 
However, these approaches rely on temporal information from previous frames, making them dependent on buffering and assumptions about frame-to-frame consistency. 
In contrast, we propose a training-free, observation-only token pruning and merging framework that operates solely on the current frame. 
By combining first-layer pruning with deep-layer merging, our TEAM-VLA achieves competitive performance without requiring temporal cues, additional training, or reliance on cross-frame priors.

\section{Preliminaries}
\subsection{Vision-Language-Action Models}
A typical vision-language action model $\pi_{\theta}$ generally consists of a sensory encoder, a language model backbone, and an action head \cite{vlasurvey,vlasurvey2}. The sensory encoder typically includes an image encoder, a language encoder, and a robot proprioception (state) encoder. These encoders transform the corresponding observation images $o_t$, task instructions $l$, and robot states $s$ into embeddings $E_{img}, E_{lang},E_{state}$, which are then passed to the language backbone and the action head for generating action $a$. Formally we defined a vision language action model conditioned on single observation as:
\begin{equation}
    a_t = \pi_{\theta}(a_{t} \mid o_t, l, s )
\end{equation}

\subsection{Token Pruning}
Given a sequence of tokens $T = \{\text{t}_1, \text{t}_2, \text{t}_3, \ldots, \text{t}_L\}$ as input to a language model (Transformer backbone), the token pruning technique removes a portion of tokens and retains a subset $T_{\text{pruned}} \subset T$. In both VLMs and VLAs, the self-attention mechanism introduces an $O(n^2)$ computational complexity, meaning that the overall computation increases quadratically with the number of tokens. Therefore, token pruning can significantly reduce the computational time required by the model without modifying its parameters.

\section{Methods}

In this section, we present the detailed architecture of our proposed TEAM-VLA. 
We first describe the overall architecture of TEAM-VLA, followed by the proposed context sampling and highlight expanding modules.
Finally, we introduce the action-guided merging strategy.


\subsection{Overall Framework}
As illustrated in Figure \ref{fig:main}, TEAM-VLA accelerates inference through a dual strategy of early pruning and mid-layer merging. 
After the sensory encoder extracts visual features, we first perform context pruning and similarity expansion to discard redundant background tokens before they enter the LLM backbone. 
This removes a large portion of spatially uninformative tokens at the earliest stage. 
Then, at an intermediate backbone layer, we apply our action-guided soft bipartite merging to compress representations by preserving the top-$M$ tokens most relevant to the action token. 
Unlike prior training-free methods \cite{wang2025specprune,efficientvla} that prune only in deeper layers (where token counts are already low), our dual-reduction design eliminates redundancy both before and within the backbone, enabling substantially greater speed-up without sacrificing task performance.


\subsection{Token Pruning}
\subsubsection{Motivation}
Token pruning requires efficiently retaining tokens that likely correspond to foreground regions of interest. 
A direct solution is to apply a segmentation model (Figure \ref{fig:expand}), but powerful foundation models such as Grounded-SAM \cite{ren2024grounded,liu2023grounding} are slow and often fail to localize objects solely from task instructions—making them impractical for real-time control. 
Other approaches \cite{xu2025vla,wang2025specprune} detect dynamic tokens via temporal differences, but they require buffering previous frames and inject task-specific priors, limiting generality. 
Similarity- or cross-attention–based methods \cite{kim2022learned,ye2025fit} are efficient, yet in a training-free setting they often produce overly sparse matches in the first layer (white activation spots in Figure \ref{fig:expand}) due to weak text–image alignment \cite{jiang2025better,wang2025specprune}. 
These limitations motivate us to develop a fast, training-free mechanism for identifying potential foreground tokens directly from the current frame, before feeding them into the language backbone.


\begin{figure}[t]
\vspace{-0.2cm}
  \centering
  \includegraphics[width=0.5\textwidth]{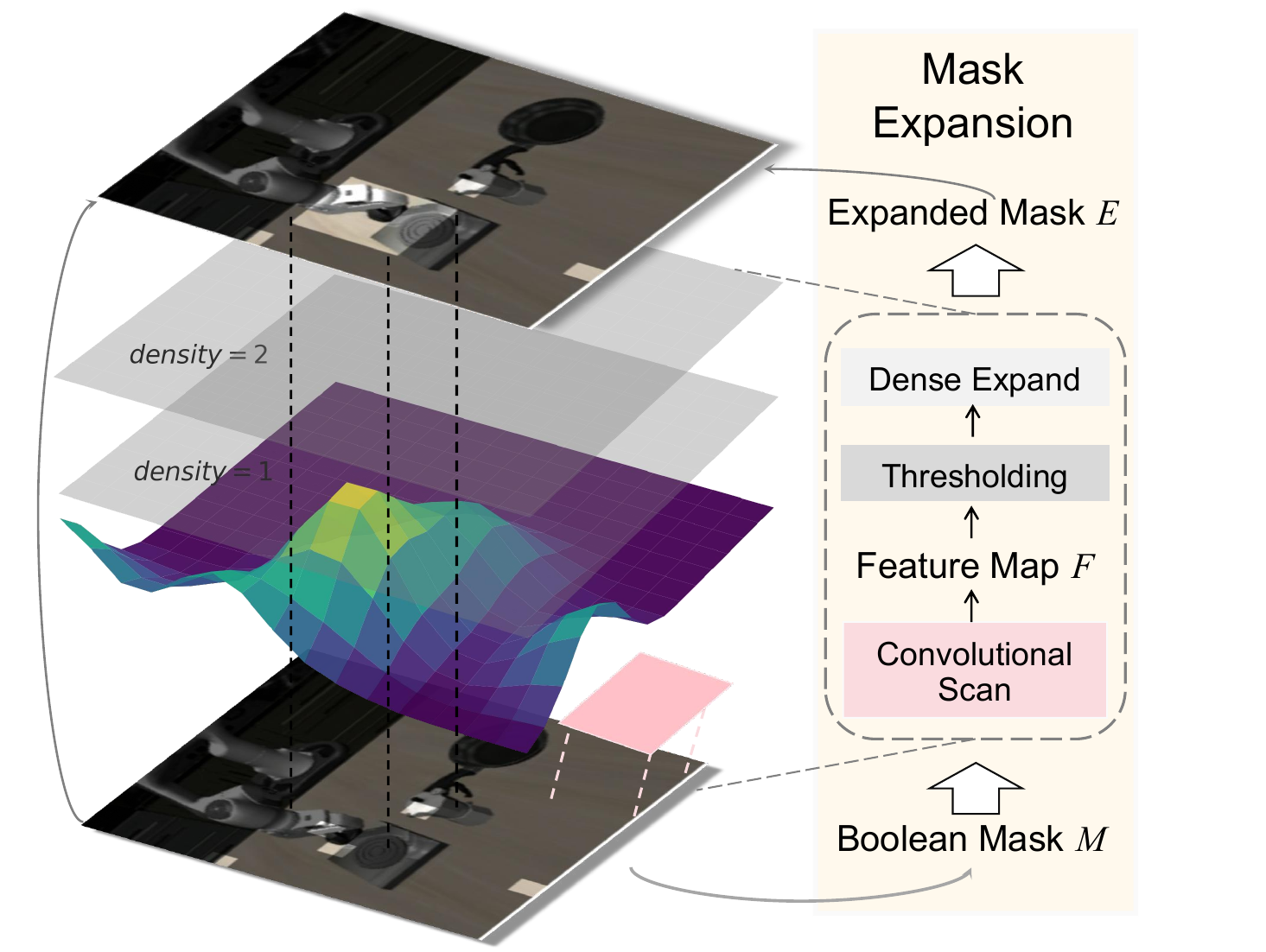} 
  \caption{We visualize the density distribution of the feature map $F$ to illustrate how attended regions aggregate spatially. On the binary mask, we apply a convolutional operation to obtain a density feature map $F$. We then expand the regions based on their density values, where the areas with the highest density are fully expanded to cover their corresponding spatial neighborhoods.}
  \label{fig:fmap}
  \vspace{-0.5cm}
\end{figure}

\subsubsection{Similarity Sampling and Token Expanding}
The goal of token expanding is to enlarge the sparse regions of task-related tokens. To achieve this, we first identify task-related sparse tokens by computing the cosine similarity between image tokens and language tokens. Formally, given a language embedding $E_{lang}$
 and image patch embedding $E_{img}$, we compute the most relevant image token for each language token using the following cosine similarity equation.
\begin{equation}
\label{eq:cossim}
\operatorname*{Argmax}_{i} \left(
    \frac{E_{\text{lang}} \cdot E_{\text{img}, i}^{\top}}
    {\|E_{\text{lang}}\|_2 \, \|E_{\text{img}, i}\|_2}
\right).
\end{equation}
Among them, we identify the patch token with the highest similarity 
$\operatorname{Argmax}_{i}$ for each language token and treat these tokens as foreground anchors in the image.
After identifying the most related image token for each language token, we binarize the resulting relevance scores to obtain a Boolean mask $M \in \{0,1\}^{p\times p}$, where $p = H(W) / s$ denotes the number of patches along the height and width and $s$ is the patch size (with $H$ and $W$ the input image height and width). To estimate the local density of attended regions, we apply a convolution operator $F = Conv(M;K)$, where $ K \in \mathbb{R}^{k \times k}$ is a kernel with all entries equal to 1 and zero padding is used. The resulting density feature map $F \in \mathbb{R}^{p \times p}$ records, for each spatial position, the number of related patches within its $k \times k$ neighborhood (shown in Figure \ref{fig:fmap}). We then update the mask via two types of regional expansion. For positions in dense neighborhoods, defined as
\begin{equation}
    \mathcal{D} = \{(i,j) \mid F_{ij} > \tau\},
\end{equation}
we assign all originally unrelated locations within the corresponding neighborhoods to 1 (related), effectively performing a deterministic dilation. 
For sparse neighborhoods,
\begin{equation}
    \mathcal{S} = \{(i,j) \mid 0 < F_{ij} < \tau\},
\end{equation}
we randomly flip one unrelated position in the local neighborhood of each $(i,j) \in \mathcal{S}$ from 0 (unrelated) to 1 (related), ensuring minimal but non-fragmented spatial coverage.
As illustrated in Figure \ref{fig:fmap}, relying solely on the sparse similarity regions (boolean mask) is insufficient to recover the full extent of the object. In contrast, our method preserves the complete task-relevant region by expanding these sparse cues into a coherent foreground representation(white plus red areas). Even with multiple image inputs, by leveraging the convolution kernel’s ability to process batch inputs (\ieno \space wrist-view and agent-view), we effectively reduce the foreground identification process to just 1–2 ms. 





\subsubsection{Context Sampling}
After identifying the foreground tokens, we randomly sample a small subset of background tokens as complementary context. 
To achieve this efficiently, we adopt a simple interval-based sampling strategy over the entire token sequence, controlled by a parameter $u \in [0,1]$. 
This ensures that a lightweight portion of scene structure is preserved, allowing the model to maintain spatial awareness while keeping redundancy minimal. 


\subsection{Token Merging}
\subsubsection{Motivation}
In VLAs, token merging has been largely overlooked \cite{vlasurvey}, despite being widely adopted in VLMs and transformer-based vision models \cite{merging,merging1,merging2}. 
We argue that while early-layer pruning effectively removes redundant spatial information and sharpens focus on the foreground, the subsequent stages should prioritize preserving this extracted foreground information—even when further reducing token count \cite{shi2025memoryvla}. 
To mitigate information loss during this process, we introduce an action-guided token merging strategy that selectively consolidates tokens based on their relevance to the action semantics.


\subsubsection{Task-Guided Bipartite Merging}
Token merging typically divides the image token set $IT$ into a source set $S$ and a target set $T$, where $|S| + |T| = |IT|$ and $S \cap T = \emptyset$.
In our method, we leverage both text and action tokens to identify the top-$M$ most relevant image tokens as the source set $S$. 
This design ensures that, unlike random or sequential selection strategies, the merging process preserves the most action-critical visual information.
Following the same similarity computation as in Equation \ref{eq:cossim}, we extract the top-$M$ tokens to form $S$, while the remaining tokens constitute $T$. 
We then perform source–target merging via a soft bipartite matching scheme. 
Concretely, we compute a similarity matrix between $S$ and $T$, which determines how each target token contributes to its closest source token during the weighted merging process. 
This preserves semantic structure while achieving efficient token reduction.

\begin{equation}
    \mathbf{Sim}
    = 
    \frac{
        \operatorname{RMSNorm}(\mathbf{S}) \times
        \operatorname{RMSNorm}(\mathbf{T})^{\top}
    }{\sqrt{d}},
    \label{eq:sim_matrix}
\end{equation}
where $\operatorname{RMSNorm}$ denotes Root Mean Square Layer Normalization, and $d$ represents the hidden dimension of the language backbone. For each target token, a probability distribution over all source tokens is obtained by applying a softmax along the source dimension:
\begin{equation}
\begin{aligned}
\mathbf{W} &= \operatorname{softmax}(\mathbf{Sim}, \text{dim}=-1), \\
W_{ij} &= \frac{\exp(Sim_{ij})}{\sum_{k=1}^{N_{Sim}}\exp(Sim_{ik})}.
\end{aligned}
\label{eq:weights}
\end{equation}
This defines a \emph{soft matching matrix} $\mathbf{W} \in \mathbb{R}^{N_{T} \times N_{S}}$. The matched target representations are aggregated back to each source position:
\begin{equation}
    \mathbf{A}
    =
    \mathbf{W}^{\top} T,
    \quad
    A_j = \sum_{i=1}^{N_{T}} W_{ij} T_{t,i},
    \label{eq:aggregation}
\end{equation}
resulting in $\mathbf{A}\in\mathbb{R}^{N_S\times d}$. We compute the total soft-matching weight received by each source token:
\begin{equation}
    \mathbf{s}
    =
    \mathbf{W}^{\top}\mathbf{1}_{N_t},
    \quad
    s_j = \sum_{i=1}^{N_{T}} W_{ij}.
    \label{eq:sum_weight}
\end{equation}
Each source token is updated by adding the aggregated target features and normalized by its total soft weight:
\begin{equation}
    S'
    =
    \frac{
        S + \mathbf{A}
    }{
        1 + \mathbf{|S|}
    },
    \label{eq:merged_source}
\end{equation}
where the division is element-wise.
Note that each target token is merged into its most similar source token, ensuring that semantically related information is preserved and preventing major information loss even under an aggressive token-reduction regime.

\begin{table*}[]
\caption{Experiments on the LIBERO Benchmark. Here, LM-pre denotes pruning applied before the large language model (LLM) backbone, and buffer-free indicates that no historical frame information is used for comparison.}
\centering
\label{tab:benchmark}
\begin{tabular}{l|cccc|ccc|cc}
\toprule
\multirow{2}{*}{Method} 
& \multicolumn{4}{c|}{Success Rate (\%) / Latency (ms)}
& \multirow{2}{*}{%
    \begin{tabular}[c]{@{}c@{}c@{}}
        Average      & \multirow{2}{*}{\raisebox{-0.5ex}{\scalebox{1}[2.0]{\space$\uparrow$}}} \\
        Success Rate & 
    \end{tabular}
}
& \multirow{2}{*}{%
    \begin{tabular}[c]{@{}c@{}c@{}}
        Average & \multirow{2}{*}{\raisebox{-0.5ex}{\scalebox{1}[2.0]{\space $\downarrow$}}} \\
        Latency &
    \end{tabular}
}
& \multirow{2}{*}{FLOPS}
& \multirow{2}{*}{LM-Pre}
& \multirow{2}{*}{Buffer-Free} \\ \cline{2-5}
& spatial & object & goal & long & & & & & \\ \hline

OpenVLA-OFT       & 97.6 / 109 & 96.5 / 109 & 97.9 / 109 & 94.5 / 109 & 96.6 & 109 & 100\% & \cmark & \cmark \\
SparseVLM         & 96.8 / 85.3 & 94.2 / 85.3 & 97.6 / 85.3 & 93.6 / 85.3 & 95.5 & 85.3 & 77\%  & \cmark & \cmark \\
VLA-Cache         & 99.0 / 101 & 97.7 / 102 & 97.4 / 102 & 93.6 / 102 & \textbf{96.9} & 101.7 & 83\% & \cmark & \xmark \\
EfficientVLA      & 96.5 / 68.8 & 91.1 / 71.4 & 96.0 / 73.7 & 72.1 / 68.6 & 88.9 & \textbf{70.6} & 35\% & \xmark & \cmark \\
SpecPrune-VLA     & 98.2 / 72.4 & 96.3 / 76.2 & 97.7 / 73.6 & 94.0 / 78.1 & 96.5 & 75.1 & 43\% & \xmark & \xmark \\ 

TeamVLA(Ours)     & 99.2 / 68.1 & 96.5 / 74.7 & 97.0 / 72.9 & 93.8 / 72.8 & {\ul 96.6} & {\ul 72.1} & 39\% & \cmark & \cmark \\
\bottomrule
\end{tabular}
\end{table*}

\begin{table}[h]
\caption{Ablation Study on LIBERO-10. Pruned means the number of tokens our method pruned. $u$ denotes the proportion of context tokens to be retained, $K$ represents the kernel size used when computing the density map via convolution.}
\label{tab:main_ablation}
\centering
\begin{tabular}{l|ccc}
\toprule
Method                                             & SR                          & \multicolumn{1}{c}{Latency} & \multicolumn{1}{c}{Pruned} \\ \hline
OpenVLA-OFT                                        & 94.5                        & 109                          & 0                          \\
\rowcolor[HTML]{FFFFFF} 
{\color[HTML]{C0C0C0} + Context Sampling ($u$ =0.1)}  & {\color[HTML]{C0C0C0} 80.1} & {\color[HTML]{C0C0C0} 67.4}  & {\color[HTML]{C0C0C0} 461} \\
\rowcolor[HTML]{ECF4FF} 
+ Context Sampling ($u$ =0.25)                        & 88.6                        & 71.3                         & 384                        \\
\rowcolor[HTML]{FFFFFF} 
{\color[HTML]{C0C0C0} + Context Sampling ($u$ =0.3)}  & {\color[HTML]{C0C0C0} 91.7} & {\color[HTML]{C0C0C0} 71.9}  & {\color[HTML]{C0C0C0} 358} \\
\rowcolor[HTML]{FFFFFF} 
{\color[HTML]{C0C0C0} + Context Sampling ($u$ =0.4)}  & {\color[HTML]{C0C0C0} 92.1} & {\color[HTML]{C0C0C0} 76.9}  & {\color[HTML]{C0C0C0} 307} \\
\rowcolor[HTML]{FFFFFF} 
{\color[HTML]{C0C0C0} + Most Similar Token}        & {\color[HTML]{C0C0C0} 89.5} & {\color[HTML]{C0C0C0} 71.9}  & {\color[HTML]{C0C0C0} 358} \\
\rowcolor[HTML]{ECF4FF} 
+ Token Expanding ($K$ size=3)                        & 93.8                        & 76.8                         & 302                        \\
\rowcolor[HTML]{FFFFFF} 
{\color[HTML]{9B9B9B} + Token Expanding ($K$ size=5)} & {\color[HTML]{9B9B9B} 94.1} & {\color[HTML]{9B9B9B} 82.5}  & {\color[HTML]{9B9B9B} 249} \\

\rowcolor[HTML]{FFFFFF}
{\color[HTML]{9B9B9B} + Action-Gudied Merging (top-50)}  
& {\color[HTML]{9B9B9B} 88.3} 
& {\color[HTML]{9B9B9B} 71.6} 
& {\color[HTML]{9B9B9B} 462} \\

\rowcolor[HTML]{ECF4FF}
+ Action-Gudied Merging(top-80)  
& 93.8 
& 72.8 
& 432 \\

\rowcolor[HTML]{FFFFFF}
{\color[HTML]{9B9B9B} + Action-Gudied Merging (top-110)}  
& {\color[HTML]{9B9B9B} 92.8} 
& {\color[HTML]{9B9B9B} 73.7} 
& {\color[HTML]{9B9B9B} 402} \\

\rowcolor[HTML]{FFFFFF}
{\color[HTML]{9B9B9B} + Action-Gudied Merging (top-130)}  
& {\color[HTML]{9B9B9B} 93.8} 
& {\color[HTML]{9B9B9B} 74.1} 
& {\color[HTML]{9B9B9B} 372} \\

\bottomrule
\end{tabular}
\end{table}

\begin{table}[h]
\caption{Ablation study on the choice of layers and Pruning vs Merging (Final top-80).}
\label{tab:layer}
\centering
\begin{tabular}{c|cc|cc}
\toprule
Method   & \multicolumn{2}{c|}{Token Merging} & \multicolumn{2}{c}{Token Pruning} \\ 
Layer    & SR             & Latency          & SR             & Latency          \\
\hline
Layer 1  & 72.6           & \textbf{68.7}             & 74.2           & \textbf{68.7}             \\
Layer 4  & 88.0           & 70.2             & 90.2           & 70.2             \\
Layer 8  & 90.4           & 71.5             & 92.2           & 71.5             \\
Layer 16 & \textbf{93.8 }          & 72.8             & 92.1           & 72.8             \\ 
\bottomrule

\end{tabular}
\end{table}
\section{Experiment}
\begin{figure*}[t]
\centering
\scalebox{0.85}{
\begin{minipage}{\textwidth}
    \subfloat[]{\includegraphics[width=0.23\textwidth]{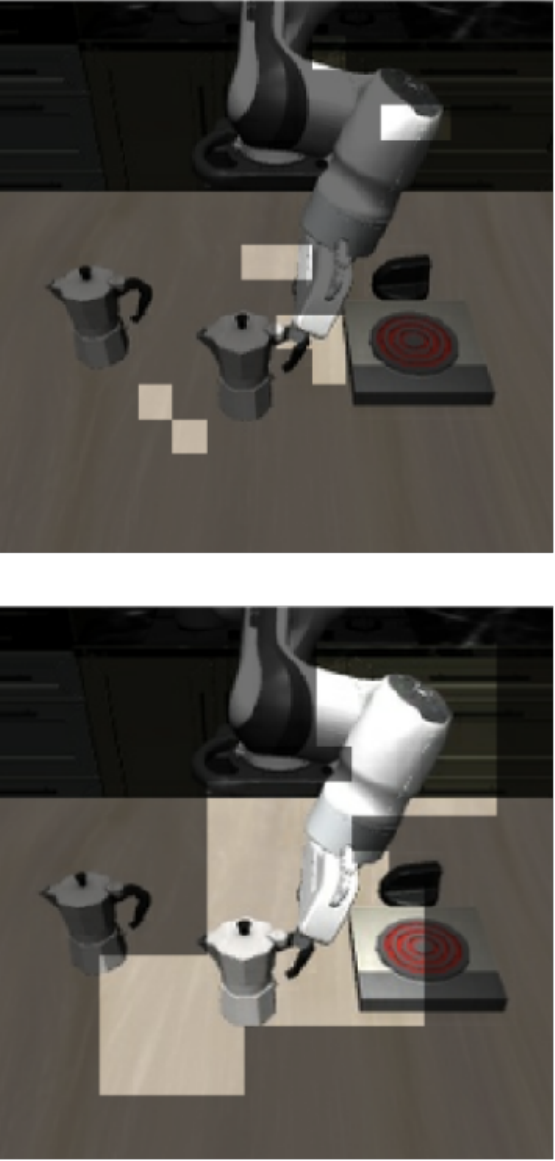}}%
    \hspace{6pt}
    \subfloat[]{\includegraphics[width=0.23\textwidth]{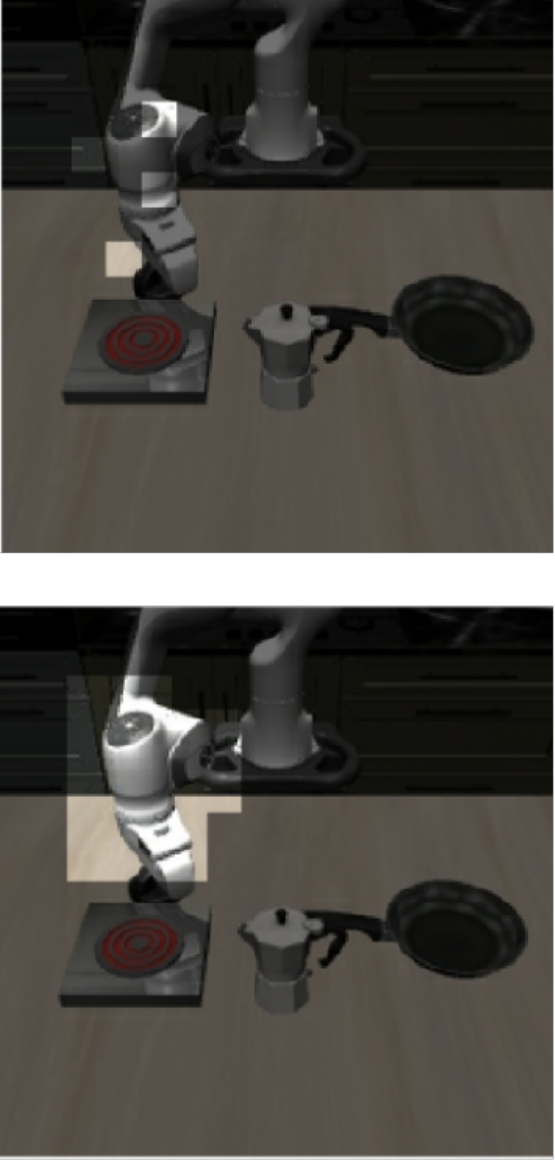}}%
    \hspace{6pt}
    \subfloat[]{\includegraphics[width=0.23\textwidth]{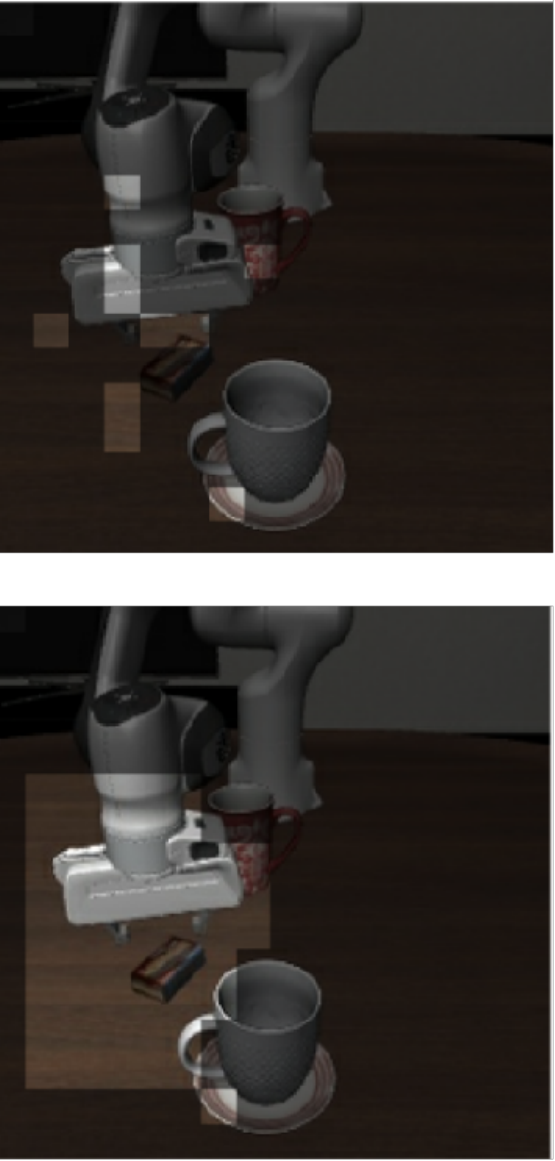}}%
    \hspace{6pt}
    \subfloat[]{\includegraphics[width=0.23\textwidth]{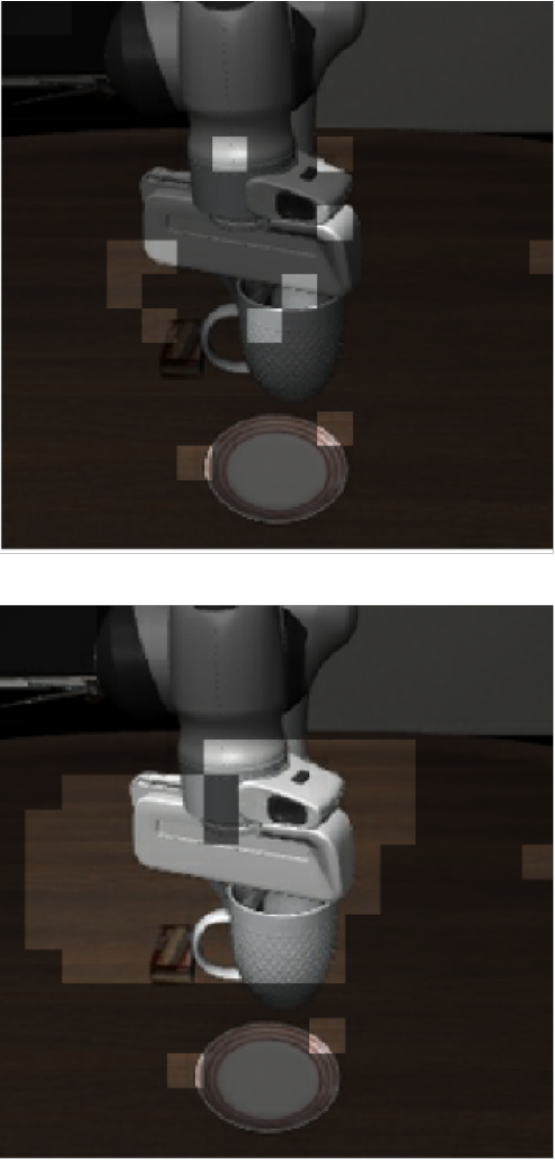}}
\end{minipage}
}
\caption{Panels (a)–(d) show visualizations of different tasks in the LIBERO-10 benchmark. For each task, the top row displays the sparse similarity mask, while the bottom row presents the corresponding expanded mask.}
\label{fig:four}
\end{figure*}
\subsection{Benchmark}
In our experiments, we use the LIBERO benchmark \cite{libero}, which comprises four task subsets: LIBERO-Spatial, LIBERO-Object, LIBERO-Goal, and LIBERO-Long.
LIBERO-Spatial evaluates spatial reasoning by requiring the robot to place a bowl relative to its current position.
LIBERO-Object assesses object recognition and manipulation across diverse items.
LIBERO-Goal examines procedural generalization by varying task goals while keeping the object set fixed.
Finally, LIBERO-Long includes ten long-horizon tasks that test the robot’s ability to execute extended, temporally complex operations.


\subsection{Experiment Setup}
\subsubsection{Compared Methods}
We compare TEAM-VLA with several state-of-the-art token-level efficiency methods designed for VLA models built on large VLM backbones. 
Specifically, we evaluate against VLA-Cache \cite{xu2025vla}, EfficientVLA \cite{efficientvla}, SparseVLM \cite{zhang2024sparsevlm}, and SpecPrune-VLA \cite{wang2025specprune}.
VLA-Cache caches static visual tokens to reduce redundant computation; EfficientVLA prunes visual tokens to accelerate action decoding; and SpecPrune-VLA removes tokens deemed irrelevant under global–local attention.

\subsubsection{Evaluation Metrics}
We adopt three widely used evaluation metrics \cite{jiang2025better,efficientvla,kim2022learned,liu2025ttf} to systematically assess our proposed TEAM-VLA: Success Rate (SR), FLOPs, and CUDA latency. 
We use success rate to evaluate task performance, FLOPs to measure theoretical computation, and CUDA latency to capture actual GPU runtime. 
We report the average CUDA latency over the ten tasks within each task suite. 

\subsubsection{Implementation Details}
Following previous token acceleration methods, we implement our proposed TEAM-VLA on OpenVLA-OFT \cite{openvla-oft}. Specifically, we utilize the codebase of LightVLA \cite{jiang2025better}. We re-ran OpenVLA-OFT and other baselines on our platform and observed that the success rate of OpenVLA-OFT was lower than that reported in previous studies, even after averaging across multiple random seeds \cite{jiang2025better}. However, for a fair and intuitive comparison, we report all baseline performances from SpecPrune-VLA \cite{wang2025specprune} and other baseline \cite{efficientvla,zhang2024sparsevlm}, which were conducted on similar devices with same CUDA latency and higher performance than our implementation. In the parameter settings, due to varying task difficulties, we apply different final merging token counts and uniform sampling rates. Specifically, we set context sample ratio $u=0.1$ and source merging token $m = 50$ for LIBERO-Spatial; $u=0.25, m = 80$ for Goal and Long; $u=0.35, m = 130$ for LIBERO-Object. The kernel size and threshold for token expanding are set to 3 and 2, respectively.
All experiments are conducted on a single NVIDIA A100-40GB GPU.

\subsection{Main Result}
Table \ref{tab:benchmark} shows that our proposed TEAM-VLA achieves competitive results compared to other state-of-the-art methods. Without any average performance degradation, TEAM-VLA reduces the inference time of OpenVLA-OFT from 109 ms to 72.1 ms, achieving over a 1.5× speedup. 
Although VLA-Cache achieves better success rate performance, its improvement on latency is marginal. 
When compared to EfficientVLA, with only an additional 1.5 ms of inference time, our method yields a 7.7\% higher success rate. Moreover, unlike VLA-Cache and SpecPrune-VLA, our method purely relies on the current observation, eliminating the need for prior knowledge derived from multiple previous frames. 
We also compare different pruning locations (before language model and in the middle of the language model) and examine whether historical frame information is necessary at the last two columns of Table \ref{tab:benchmark}. For pruning location, existing attention-based methods typically rely on the second or third layer of the language backbone, or even deeper layers, where action tokens, language tokens, and image tokens are better aligned \cite{wang2025specprune,efficientvla}. However, pruning at such depths fails to deliver meaningful speed gains due to the significant computation already incurred in earlier layers \cite{wang2025specprune}. 

In addition, we evaluate the use of an image buffer. Previous work\cite{liu2025ttf,wang2025specprune,xu2025vla} identifies foreground regions by comparing patches between consecutive frames. This patch-based frame differencing method is motivated by the fact that, in robotic manipulation tasks, the moving entities are typically the robot arm or the objects being manipulated. Although such temporal differencing can quickly localize foreground regions, it introduces the need to store historical frames and relies on temporal priors, making the method less flexible. In contrast, our buffer-free strategy achieves comparable effectiveness without relying on any historical observations.


\begin{figure}[h]
  \centering
  \setlength{\abovecaptionskip}{-0.1cm}
  \includegraphics[width=0.5\textwidth]{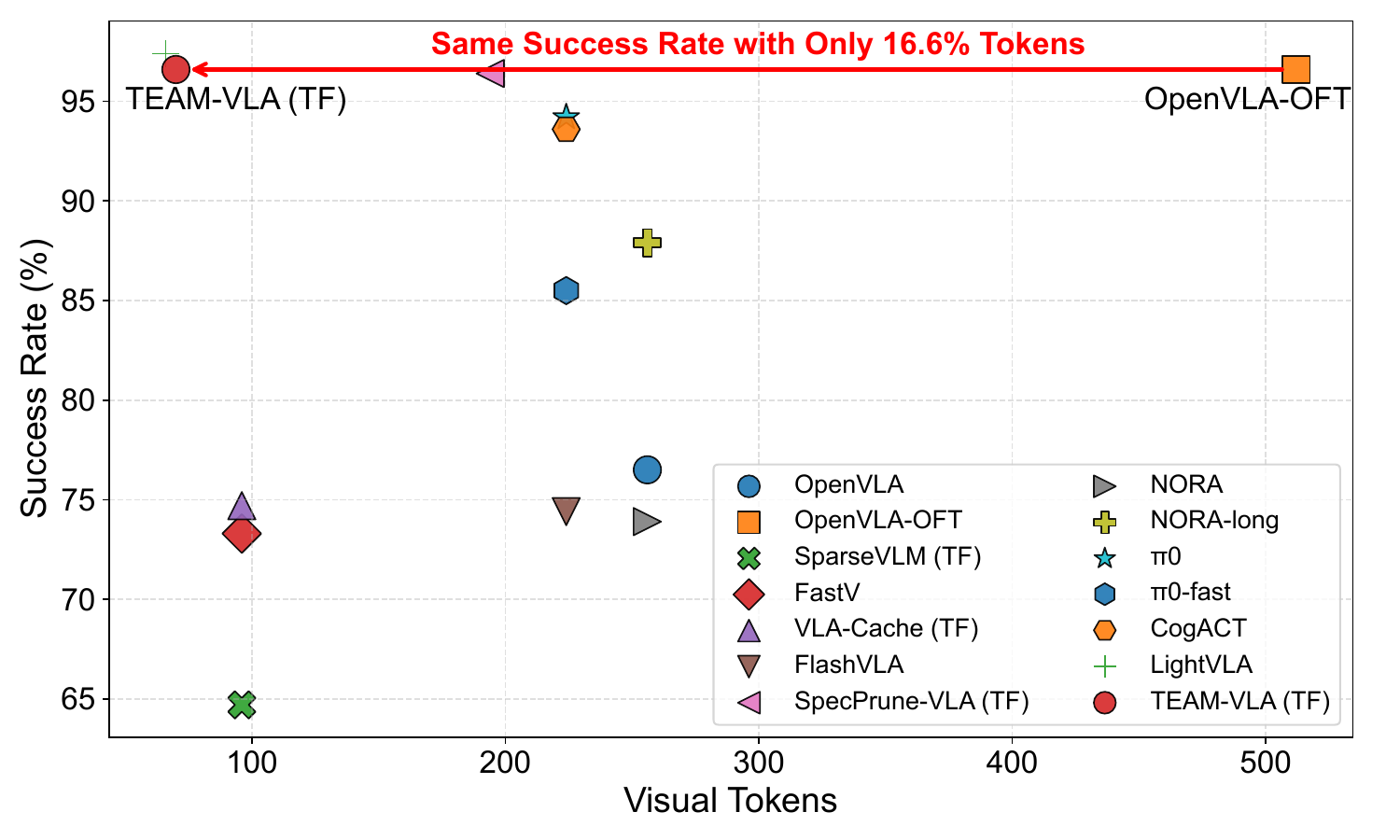} 
  \caption{We further compare the number of remaining tokens across several mainstream methods, reporting for TEAM-VLA the average token count after the merging stage. As shown, TEAM-VLA retains substantially fewer tokens than other training-free (TF) approaches while achieving significantly higher performance.}
  \label{fig:duibi}
\end{figure}

\begin{table*}[t]
\caption{Ablation Study with Threshold  $\tau$.}
\label{tab:tau}
\centering
\begin{tabular}{c|cccc|ccc}
\toprule
\multirow{3}{*}{\begin{tabular}[c]{@{}c@{}}Thres. \\ $\tau$\end{tabular}}
& \multicolumn{4}{c|}{Success Rate (\%) / Latency (ms) / Patches Number}
& \multirow{3}{*}{\begin{tabular}[c]{@{}c@{}}Avg \\ Succ.\end{tabular}}
& \multirow{3}{*}{\begin{tabular}[c]{@{}c@{}}Avg \\ Latency\end{tabular}}
& \multirow{3}{*}{\begin{tabular}[c]{@{}c@{}}Avg \\ Patch\end{tabular}} \\ \cline{2-5}
& \begin{tabular}[c]{@{}c@{}}spatial\\ ($u=0.1$)\end{tabular}
& \begin{tabular}[c]{@{}c@{}}object\\ ($u=0.35$)\end{tabular}
& \begin{tabular}[c]{@{}c@{}}goal\\ ($u=0.25$)\end{tabular}
& \begin{tabular}[c]{@{}c@{}}long\\ ($u=0.25$)\end{tabular}
& & & \\ \hline
$\tau = 0$ &
98.4 / 85.1 / 336 &
97.0 / 88.5 / 375 &
97.6 / 85.4 / 323 &
95.6 / 84.9 / 335 &
97.2 & 86.0 & 342.3 \\
$\tau = 1$ &
99.2 / 69.6 / 151 &
96.6 / 78.2 / 245 &
96.8 / 76.0 / 187 &
93.8 / 77.2 / 201 &
96.6 & 75.3 & 196.0 \\
$\tau = 2$ &
96.6 / 68.8 / 92 &
96.4 / 77.2 / 202 &
95.2 / 70.0 / 150 &
90.0 / 71.1 / 160 &
94.6 & 71.8 & 151.0 \\
\bottomrule
\end{tabular}
\end{table*}

\subsection{Ablation Study}
\subsubsection{Ablation Study on Components}
Table \ref{tab:main_ablation} illustrates the effect of TEAM-VLA’s components. First, we evaluate the proposed Token Expanding method by comparing context sampling ($u=0.4$) with Token Expanding ($u=0.25, K=3$). We find that, with a similar number of pruned tokens, our proposed method yields a 1.7\% improvement in success rate. We also compare the performance without Token Expanding. When only language-related tokens are added, the model yields suboptimal performance with a success rate of 89.5\%, which affirms the importance of Token Expanding when pruning in the first layer. We further study the effect of the kernel size in Token Expanding. With larger kernel size ($K=5$), although the success rate increases by 0.3\%, the dense area expands to a larger spatial region, resulting in more visual tokens during inference and an additional 6 ms of latency.

\subsubsection{Ablation Study on Merging Layer}
In the first-stage token pruning, TEAM-VLA performs an initial pruning step after the sensory encoder processes the raw input but before the tokens are fed into the language backbone. Prior work \cite{jiang2025better} has shown that applying pruning directly at the sensory encoding stage can degrade the model’s visual representation quality, leading to suboptimal performance. Conversely, performing pruning only within the early layers of the language backbone yields limited speedup \cite{efficientvla,wang2025specprune}. Therefore, we conduct an ablation study on the choice of layer at which merging is performed. As shown in the Token Merging column of Table \ref{tab:layer}, the success rate (SR) consistently increases when merging is performed at deeper layers, which aligns with findings from prior work. Since merging serves as the second-stage acceleration in our framework, we select the middle layer, where the highest success rate is achieved, as the merging location in our experiments.
\subsubsection{Avlation Study on Threshold $\tau$}
From the ablation results of Table \ref{tab:tau}, increasing the density threshold $\tau$ reduces the number of expanded regions, resulting in fewer patches and lower latency. For example, the average patch count decreases from $342 \rightarrow 196 \rightarrow 151$ when $\tau$ increases from 0 to 1 to 2, and the corresponding average latency drops from$ 86.0\text{ms}  \rightarrow 75.3 \text{ms} \rightarrow 71.8 \text{ms}$. However, a large $\tau$ also reduces useful context, causing the success rate to fall from 97.2\% ($\tau=0$) to 94.6\% ($\tau=2$). In contrast, $\tau=1$ maintains high accuracy (96.6\%) while significantly reducing computation. Therefore, we adopt $\tau =1 $ as the default setting for our main experiments.

\subsection{Further Analysis}

\subsubsection{Merging vs Pruning}
Table \ref{tab:layer} further illustrates the difference between merging and pruning. 
Although the performance gap is modest, merging at deeper layers consistently outperforms pruning. 
This is because, after the initial pruning stage, most remaining tokens already correspond to foreground or otherwise informative regions. 
Directly pruning these tokens down to a small budget (\egno, 80) inevitably removes valuable information. 
In contrast, merging aggregates less critical tokens into the most relevant ones, preserving their semantic content while reducing token count.


\subsubsection{Final Token Comparison}
 The last three rows of Table \ref{tab:main_ablation} present the effect of the final number of merged tokens. When the number of tokens is reduced to 50, the success rate drops to 88.3\%. However, increasing the token count to 80 leads to a substantial improvement, achieving a success rate of 93.8\%. Further increasing the number of tokens yields no additional performance gains. These results demonstrate that our merging strategy effectively preserves the information carried by tokens that would otherwise be discarded. As a result, proposed action-guided merging maintains comparable success rates to the pre-merging model while simultaneously improving inference speed. We further compare the final number of retained tokens across different methods in Figure \ref{fig:duibi}. Among all Training-Free (TF) approaches, our method significantly outperforms the others, achieving substantially fewer final tokens while maintaining comparable performance. Compared to the baseline, TEAM-VLA removes the majority of tokens, on average 432, yet still preserves the same level of task success.
 
\subsubsection{Visualization of Expanded}
We provide a visualization of the extracted patches in Figure \ref{fig:four}, where the first row shows the sparse image–text most similar patches, and the second row presents the patches after our expansion. As illustrated, the most similar patches contain cues about the objects and the robot arm, but their high sparsity prevents the model from capturing complete structural information about the arm or the surrounding spatial context. Our expansion method reconstructs a more coherent and spatially continuous local foreground region, thereby enabling the model to access richer visual information and ultimately leading to improved performance.

\section{Conclusion}
In this paper, we present TEAM-VLA, a training-free framework that efficiently identifies foreground tokens and performs action-guided merging. 
Unlike prior methods, TEAM-VLA requires no prior knowledge or temporal cues; instead, it uses a density-expansion strategy to reconstruct object foreground regions and the robot arm from sparse similarity signals. 
A mid-layer merging module further accelerates inference while preserving essential semantics. 
Extensive experiments show that TEAM-VLA achieves state-of-the-art performance among training-free approaches, delivering strong accuracy–efficiency trade-offs for VLA models.


\bibliographystyle{IEEEtran}
\bibliography{main}


\end{document}